\documentclass[10pt,twocolumn,letterpaper]{article}
\usepackage{iccv}
\usepackage{mycommands}

\iccvfinalcopy 

\def\iccvPaperID{1} 

\ificcvfinal\fi

\begin{document}

\title{Semantic Cross-View Matching}
\author{Francesco Castaldo\textsuperscript{1}  \enspace Amir Zamir\textsuperscript{2}  \enspace Roland Angst\textsuperscript{2,3}  \enspace Francesco Palmieri\textsuperscript{1}  \enspace Silvio Savarese\textsuperscript{2}\\
\textsuperscript{1}Seconda Universit\`a di Napoli \\ \textsuperscript{2}Stanford University  \\ \textsuperscript{3}Max Planck Institute for Informatics\\
{\tt\small \{francesco.castaldo,francesco.palmieri\}@unina2.it  \{zamir,ssilvio\}@stanford.edu rangst@mpi-inf.mpg.de}
}
\maketitle

\begin{abstract}
Matching cross-view images is challenging because the appearance and viewpoints are significantly different. While low-level features based on gradient orientations or filter responses can drastically vary with such changes in viewpoint, semantic information of images however shows an invariant characteristic in this respect.
Consequently, semantically labeled regions can be used for performing cross-view matching.

In this paper, we therefore explore this idea and propose an automatic method for detecting and representing the semantic information of an RGB image with the goal of performing cross-view matching with a (non-RGB) geographic information system (GIS).
A segmented image forms the input to our system with segments assigned to semantic concepts such as traffic signs, lakes, roads, foliage, etc.
We design a descriptor to robustly capture both, the presence of semantic concepts and the spatial layout of those segments. 
Pairwise distances between the descriptors extracted from the GIS map and the query image are then used to generate a shortlist of
the most promising locations with similar semantic concepts in a consistent spatial layout. An experimental evaluation with challenging query images and a large urban area shows promising results.
\end{abstract}

\section{Introduction} \label{sec:introduction}
In this paper, we consider the cross-view and cross-modality matching problem between street-level RGB images and a geographic information system (GIS). Specifically, given an image taken from street-level, the goal is to query a database assembled from a GIS in order to return likely locations of the street-level query image which contain similar semantic concepts in a consistent layout. 
Relying only on visual data is important in GPS-denied environments, for images where such tags have been removed on purpose (e.g. for applications in intelligence or forensic sciences), for historical images, or images from the web which are lacking any GPS tags.
Traditionally, such matching problems are solved by establishing pairwise correspondences between interest points using local descriptors such as SIFT~\cite{lowe2004} with a subsequent geometric verification stage. Unfortunately, even if top-down satellite imagery is available, such an approach based on local appearance features is not applicable to the wide-baseline cross-view matching considered in our setting, mainly because of the following two reasons. 
Firstly, the extremely wide baseline between top-view GIS imagery and the street-level image leads to a strong perspective distortion, and secondly, there can be drastic changes in appearance, e.g. due to different weather conditions, time of day, camera response function, etc.

In this paper, we present a system to handle those two challenges. We propose to phrase the cross-view matching problem in a semantic way. Our system makes use of two cues: what objects are seen and what their geometric arrangement is. This is very similar to the way we humans try to localize ourselves on a map. For instance, we identify that a house can be seen on the left of a lake and that there are two streets crossing in front of this house. Then we will look for the same semantic concepts in a consistent spatial configuration in the map to find our potential locations. 
\begin{figure}[tb]
 \centering   
\includegraphics[width=0.47\textwidth, keepaspectratio=true]{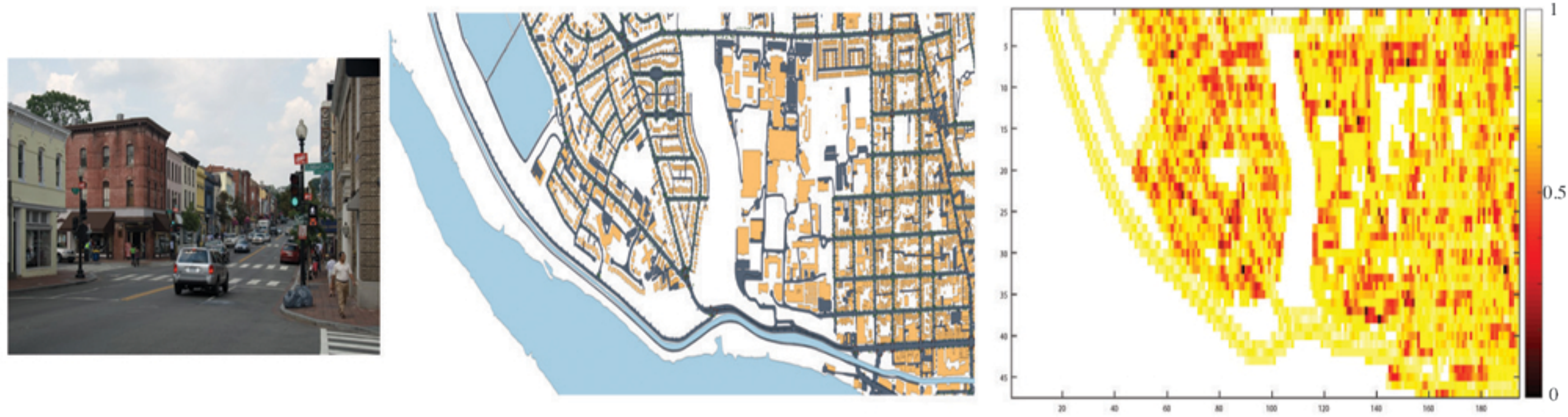}
 \caption{Typical examples for a query image (left), semantic map from the GIS (middle) and corresponding heat map (right). Our proposed geolocalization scheme captures the semantic layout of the query image and leverages it for matching to a semantic map.}
  \label{fig:exampleimages}
\end{figure}
Inspired by this analogy, in our system, instead of matching low-level appearance-based features, we propose to extract segments from the image and label them with a semantic concept employing imperfect classifiers which are trained using images of the same viewpoint and therefore are not invalidated by the viewpoint change.
GIS often already provide highly-accurate semantically annotated top-down views thereby rendering the semantic labeling superfluous for the GIS satellite imagery. Hence, we assume that such a semantic map is provided by the GIS. A typical query image and an excerpt of a semantic map can be seen in \figref{fig:exampleimages}. 
%
%
The semantic concepts we focus on (e.g., buildings, lakes, roads, etc) form large (and quite often insignificant in number) segments in the image, and not points. Therefore, we argue that a precise point-based geometric verification, like a RANSAC-search\cite{fischler1981random} with an inlier criterion based on the Euclidean distance between corresponding points, is not applicable. 
We address these issues by
designing 
a descriptor to robustly capture the spatial layout of those semantic segments. Pairwise asymmetric L2 matching between these descriptors is then used to find likely locations in the GIS map with a spatial layout of semantic segments which is consistent with the one in the query image. We also develop a tree-based search method based on a hierarchical semantic tree to allow fast geo-localization in a geographically broad areas.

\begin{figure*}[htb]
 \centering   
\includegraphics[width=\textwidth]{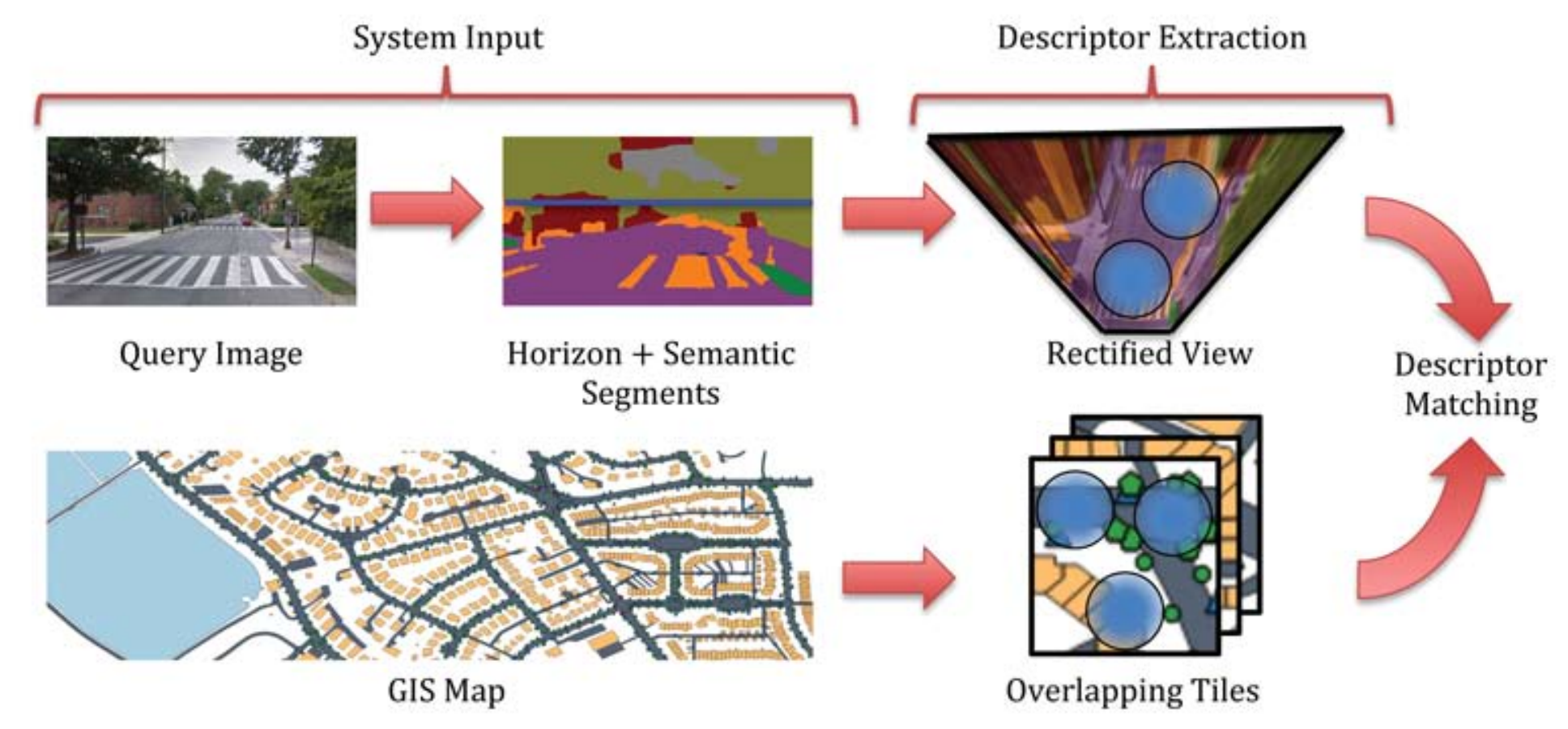}
 \caption{
 Pipeline of our system: We use existing methods to compute the system inputs, namely the vanishing line of the ground plane and a semantic segmentation of the image. The ground plane in the query image is rectified to minimize distortions between a top-down GIS map and the query image. The GIS map is split into multiple overlapping tiles. Descriptors (blue circles) which capture the layout of the semantic regions in the query image and the tiles are extracted. Pairwise distances between those descriptors are then used for a final ranking of promising tiles.
}
  \label{fig:systemoverview}
\end{figure*}

\section{Related Work} \label{sec:relatedwork}
Cross-view matching in terms of semantic segments between street-level query image and a GIS map joins several previous research directions. 
Matching across a wide baseline has traditionally been addressed with local image descriptors for points~\cite{lowe2004}, areas~\cite{matas2004robust}, or lines~\cite{bay2005wide,verhagen2014scale}.
Registration of street-level images with oblique aerial or satellite imagery is generally based only on geometric reasoning. Previous work, e.g. \cite{frueh2003constructing,kaminsky2009alignment}, has reduced the matching problem to a 2D-2D registration problem by projecting ground models along vertical directions and rectifying the ground plane. Unlike our approach, the mentioned work requires a $3$D point cloud at query time, either from a laser scan~\cite{frueh2003constructing} or from multiple views based on structure-from-motion~\cite{kaminsky2009alignment}.
More recently, \cite{shan2014accurate} considered the registration problem of a dense multi-view-stereo reconstruction from street-level images to oblique aerial views. 
Building upon accurate 3D city models for assembling a database, contours of skylines in an upward pointing camera can be matched to a city model~\cite{ramalingam2010skyline2gps} or perspective distortion can be decreased by rectifying regions of the query image according to dominant scene planes~\cite{baatz2012leveraging}. \cite{bansal2012ultra} also relied on rectification of building facades, however, their system relied on the repetitive structure of elements in large facades, enabling a rectification without access to a 3D city model.
Using contours between the sky and landscape has also been shown to provide valuable geometric cues when matching to a digital elevation model~\cite{baatz2012large}.
Not using any 3D information, Lin et al.~\cite{lin2013cross} proposed a method to localize a street-level image with satellite imagery and a semantic map. Their system relies on an additional, large dataset which contains GPS-annotated street-level images which therefore establish an explicit link between street-level images and corresponding areas in the satellite imagery and semantic map. Similarly to the idea of information transfer in ExemplarSVMs~\cite{malisiewicz2011ensemble}, once a short-list of promising images from this additional dataset has been generated by matching appearance-based features, appropriate satellite and semantic map information can be transferred from this short-list to the query image.

%
%
%
Visual location recognition and image retrieval system emphasise the indexing aspect and can handle large image collections: Bag-of-visual-words~\cite{Sivic2003}, vocabulary trees~\cite{Nister} or global image descriptors such as Fisher vectors \cite{Jegou2012} have been proposed for that purpose, for example.
All those schemes do not account for any higher-level semantic information. More recently, \cite{durand2014semantic} has therefore introduced a scheme where pooling regions for local image descriptors are defined in a semantic way: detectors assign each segment a class label and a separate descriptor (e.g. a Fisher Vector) is computed for each such segment.
Those descriptors rely on local appearance features, which fail to handle significant viewpoint changes faced in the cross-view matching problem considered in our paper.
Also, this approach does not encode the spatial layout \textit{between} semantic segments.
If the descriptors are sufficiently discriminative by themselves, encoding this spatial layout is less important. In our case however, the information available in the query image which is shared with the GIS only captures class labels and a very coarse estimate of the segment shapes.
It is therefore necessary to capture both, the presence of semantic concepts  and the spatial layout between those concepts, in a joint representation.
%

Very recently, Ardeshir et al.'s work~\cite{ardeshir2014gis} considered the matching problem between street-level image and a dataset of semantic objects. Specifically,  deformable-part-models (DPMs) were trained to detect distinctive objects in urban areas from a single street-level image. The main objective of that paper was improved object detection with a geometric verification stage using a database of objects with known locations and the GPS-tag and viewing direction of the image has been assumed to be known roughly.
 They also present an exhaustive search based approach to matching an image against the entire object database. The considered DPMs in \cite{ardeshir2014gis} are well-localized and can be reduced to the centroid of the detection bounding box for a subsequent RANSAC step which searches for the best 2D-affine alignment in image space. Therefore, \cite{ardeshir2014gis} can only handle such "spot" based information and is not designed to handle less accurate and potentially larger semantic segments with uncertain locations such as the ones provided by classifiers for `road' or `lake'.

%


\section{Semantic Cross-View Matching} \label{sec:systemoverview}
A graphical illustration of our proposed system is shown in \figref{fig:systemoverview}. The building blocks proposed by our paper will be described in detail in the next section, here we provide a rough overview of the system and describe the system inputs.
The computation of this input information relies on previous work and is therefore not considered as one of our contributions.
%
Given a street-level query image, first we split it into superpixel segments and label each segment with a semantic concept by a set of pre-trained classifiers.
We train two different types of classifiers to annotate superpixels with class labels corresponding to a subset of the labels available in the GIS\footnote{GIS often provide very fine-grained class labels for regions or areas. For simplicity, we consider a subset of labels for which appearance based classifiers can be trained reliably.}. 
For semantic concepts with large variation in appearance and shape, we are using the work by Ren et al.~\cite{ren2012rgb}. However, in street level images, it is also quite common to spot highly informative objects with small with-in class variation. Typical examples are traffic signs or lamp posts. For each of those classes, we therefore employ a deformable-part-model (DPM), similar to the ones of \cite{ardeshir2014gis}.
The second piece of input information is an estimate of the vertical vanishing point or the horizon line, e.g.~\cite{Hartley2004,denis2008efficient,bazin2012globally} describe how those entities can be estimated from a single image.
The perspective distortion of the ground plane can then be undone by warping the query image with a suitable rectifying homography, which is fully determined by the horizon line, assuming a rough guess of the camera intrinsic matrix is available.
We have opted to estimate the two inputs for our system, namely the ground plane location and the superpixel segmentation, in two entirely independent steps. We note however, that at the expense of higher computational cost this could be estimated jointly~\cite{hoiem2008putting,saxena2007learning}.


The semantic map from the GIS and the warped and labelled query image now share the same modality and suffer from less perspective distortion. The cross-view matching problem between query and semantic GIS map is then cast as a search for consistent spatial layouts of those semantically labeled regions. 
\mydelete{However, special care must be taken to model the propagation of errors due to potentially inaccurately detected segments in the original image 
and due to segments which are not contained in the ground plane.
}
In order to do so,
we design a Semantic Segment Layout (SSL) descriptor which captures the \emph{spatial} and \emph{semantic} arrangement of those segments within a local support area located at a point in the query image or the semantic map.
Upon extracting such descriptors from the rectified image and semantic map, the problem can be reduced to a well-understood matching problem. 

\section{Semantic Segment Layout (SSL) Descriptor}
%
The goal of the SSL descriptor is to capture the \emph{presence} of semantic concepts and at the same time encode the \emph{rough geometric layout} of those concepts.
Similar to several previous descriptors~\cite{belongie2000shape,tola2008fast}, the neighborhood around the descriptor centre is captured with pooling regions arranged in an annular pattern where the size of the regions increases with increasing distance to the descriptor centre.
We instantiate separate pooling regions for each semantic concept and the overall descriptor is the concatenation of the per-concept descriptors.

%
%
%
%

\subsection{Placement of Descriptors}
\begin{figure*}[!htb]
 \centering   
\includegraphics[width=\textwidth]{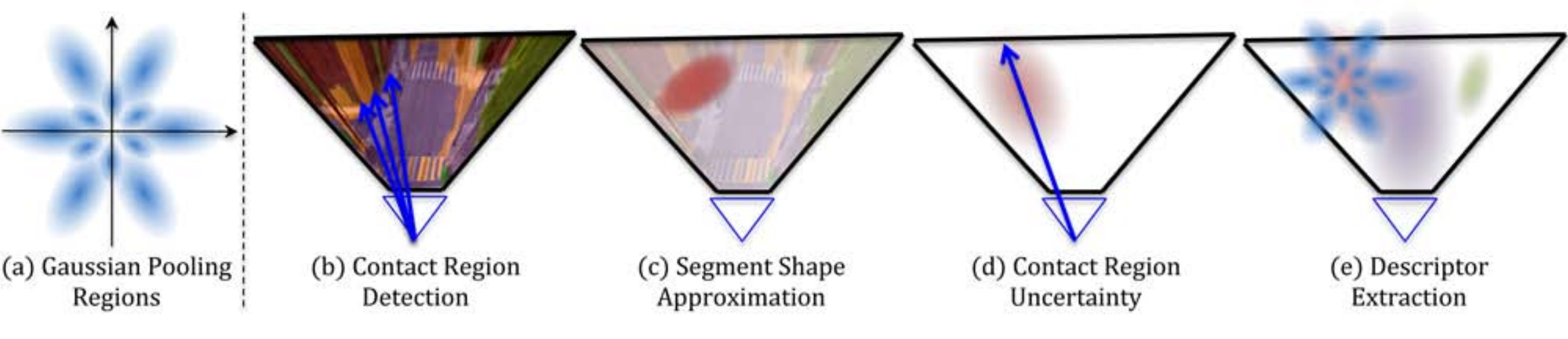}
 \caption{
Pipeline for the SSL descriptor extraction in a query image: Due to the challenges in the contact region detection, in practice a simplified version of the illustration shown here is used, see text for more details.
(a) The pooling regions of the SSL descriptor are a set of Gaussians. The descriptor visualized here contains 2 rings of six pooling regions each. (b) Contact regions of vertical objects are detected in the rectified image by shooting rays from the camera position projected onto the ground plane  (visualized as the lower vertex of the blue triangle and computed as $\mymatrix{H} \myvector{v}_{|}$, where $\mymatrix{H}$ is the rectifying homography and $\myvector{v}_{|}$ is the vertical vanishing point) and recording the closest intersection point with the segment. (c) Semantic segments are approximated with a Gaussian (for clarity, visualization is only for a single segment). (d) Uncertainty in contact regions are handled by convolving the Gaussians of vertical segments with another Gaussian whose covariance is elongated along the line between the camera position and the segment centroid. (e) A descriptor is extracted at each segment centroid.
 }
  \label{fig:descrextraction}
\end{figure*}
Similar to appearance based local descriptors, we have to choose the locations where to extract descriptors and its orientation.
We are not aware of a good way to find reliable interest points in arrangements of semantic regions.

We initially experimented with placing a separate descriptor at the center of each semantic segment. Unfortunately, this choice turns out to be very sensitive to the location of the segment projected onto the ground plane. This projection depends on an accurate estimate of the contact region of that segment with the ground plane. Based on our experiments, it is challenging to get a sufficiently accurate estimate without manual user intervention. 
In some preliminary experiments, we therefore also tried to factor in the contact region uncertainty by 'blurring' the contributions of a neighbouring segment along the line of sight between the camera and that segment, with an increasing amount of blur the further away the segment is from the camera center. We think this is an elegant and theoretically sound way to account for those uncertainties
and we refer to \figref{fig:descrextraction} for a graphical illustration of the descriptor and of the subsequent processing steps of our preliminary pipeline.
We plan to explore this pipeline in future work in more detail.
In this work however, we settled for a simpler choice: a single descriptor is placed either in the center of the rectified image (denoted CI in the experiments) or at the center of the camera (CC in the experiments). Also, despite the SSL descriptor being more general, we only choose one annular pooling region thereby putting more emphasis on capturing the direction of semantic segments rather than the direction \textit{and} distance. This choice is again motivated by the difficulty of accurately estimating contact regions.

%
We suggest using descriptors which are not rotation invariant and an orientation therefore needs to be assigned to each descriptor as well.
The reason for this choice is that the alternative of defining a rotation-invariant descriptor leads to a considerably less discriminative descriptor and pairwise matching score.
It is straightforward to define a canonical orientation in the semantic GIS map. For example, the first pooling region can be chosen to point to geographic North. However, unless compass direction is available, it is not easily possible to define a canonical orientation for the rectified query image. 

Hence, we choose an arbitrary 
direction for
the descriptors extracted from the rectified query image, and cope with not knowing the rotation parameters by employing a rotation invariant distance metric at the query time (see section \secref{sec:matching}).


\subsection{Aggregation over Pooling Region} \label{descreval}
There are several ways to capture the 'overlap' between a pooling region and a semantic segment. An intuitive approach is to compute the area of intersection between the pooling region and the segment.
This can be fairly slow for irregularly shaped segments. Moreover, the shape of the segments in the query image are quite imprecise, so an accurate computation of the intersection area might be unnecessary or even harmful.
Here, we propose a scheme motivated by a probabilistic point of view. The segments can be considered as a spatial probability distribution, that a point sampled in or close to this segment takes a certain label. Similarly, the pooling regions are interpreted as probability distributions of sampling a point at a certain location. We then have to compute a statistical measure for the 'similarity' between the two distributions. For discrete distributions, the mutual information is a good candidate. In our setting however, we have to handle continuous distributions defined over the ground plane. The Bhattacharyya distance~\cite{Abou} is a good way to measure the overlap between two continuous distributions  and can be efficiently computed when we deal with Gaussian distributions.
Hence, in practice, in order to keep the computational requirements low, we will use a two-dimensional Gaussian to define a pooling region and each segment will be approximated by a Gaussian, as well. 
In detail, let ${\cal G}^s(x;\mu^s,\Sigma^s)$ and ${\cal G}^p(x;\mu^p,\Sigma^p)$ denote the Gaussian for the segment and the pooling region, respectively. The Bhattacharyya distance is then given by
\begin{align}
d_B({\cal G}^s,{\cal G}^p) = \frac{1}{8}(\mu^s-\mu^p)^T\Sigma^{-1}(\mu^s-\mu^p)+ \nonumber \\
\frac{1}{2}\ln\frac{\det{\Sigma}}{\sqrt{\det{\Sigma^s}\det{\Sigma^p}}},
\end{align}
where $\Sigma=\frac{\Sigma^s+\Sigma^p}{2}$. 
If multiple segments that are labeled with the same semantic concept are present, they can be treated as a Gaussian Mixture Model (GMM). The Bhattacharyya distance between two GMMs can be approximated \cite{Sfikas2005} by
$
d_B({\cal M}^s,{\cal M}^p)=\sum_{i=1}^{N^s}\sum_{j=1}^{N^p}\alpha_i\beta_j d_B({\cal G}^s_i,{\cal G}^p_j)
$, 
where ${\cal M}^s=\sum_{i=1}^{N^s}\alpha_i{\cal G}^s_i(x;\mu^s_i,\Sigma^s_i)$ and ${\cal M}^p=\sum_{j=1}^{N^p}\beta_j{\cal G}^p_j(x;\mu^p_j,\Sigma^p_j)$ are a GMM for the segment and the pooling region, respectively. In our case, the pooling region is always represent by a single Gaussian, so $N^p = 1$ and $\beta_1 = 1$.
The Bhattacharyya distance is then converted to the Hellinger distance\footnote{The Hellinger distance satisfies the triangle inequality whereas the Bhattacharyya does not.} 
\begin{align}
  d_H({\cal M}^s,{\cal G}^p) &= \sqrt{1 - BC({\cal M}^s,{\cal G}^p)}
  \text{,}
\end{align}
where
$ BC({\cal M}^s,{\cal G}^p) = \exp\left(-d_B({\cal M}^s,{\cal G}^p) \right) $
is the Bhattacharyya coefficient.
Our descriptor is the concatenation of all those Hellinger distances
$ \myvector{d} = \left( d_H({\cal M}^1, {\cal G}^1), d_H({\cal M}^1, {\cal G}^2), \ldots,
d_H({\cal M}^{N_s}, {\cal G}^{N_p}) \right) \in \Re^{N_s N_p}$. 
Each block of this descriptor corresponding to a semantic concept is then L2-normalized independently of the other blocks.
If a concept is not present, the Hellinger distance for all pooling regions of that concept are set to zero.
\mydelete{Special care must be taken during the extraction of descriptors from the rectified query image, due to geometric imprecision in the localization of semantic segments. We also refer to \figref{fig:descrextraction} for a graphical illustration of the necessary processing steps.
Superpixels which are not contained in the ground plane will be mapped to a wrong location by the rectifying homography. Furthermore, those superpixels are likely not visible from the top-down view in the GIS database. However, what is often visible from the top-down view is the projection of those superpixels onto the ground plane, which defines the silhouette of the object as viewed from the top.
We therefore estimate the contact region of a vertical object with the ground plane and only aggregate this contact region in our descriptor, see \figref{fig:descrextraction}(b) for an illustrative explanation.
Unfortunately, accurate detection of the contact regions is a challenging task. We therefore need a procedure to handle inaccurate estimates.
An important observation is that the location uncertainty of the contact regions is highly anisotropic. For example, typical street-level images have the y-axis roughly aligned with the vertical vanishing point. In that case, the contact region of a tree or traffic sign is fairly well localized side-wise, however the depth uncertainty w.r.t. the camera can be large. Inaccuracies in contact point estimation therefore mostly lead to displacements in the rectified image along the line between the true contact point and the camera center projected onto the ground plane, i.e. $\mymatrix{H} \myvector{v}_{|}$, where $\mymatrix{H}$ is the rectifying homography and $\myvector{v}_{|}$ is the vertical vanishing point.
We design our descriptor to account for these inaccuracies by blurring the contribution of each semantic segment along this line with a spatially varying anisotropic blur kernel. The covariance matrix of the blur kernel is largest along the line of sight and increases with distance to the camera, see also \figref{fig:descrextraction}(d). 
}

\subsection{Descriptor Computation in Semantic Map}
The descriptor extraction from the semantic map is considerably simpler than from the query image: no rectification needs to performed as the overhead view can be readily rendered and the segment boundaries are exact. 
The semantic map is divided into overlapping tiles for the subsequent matching stage.
A single SSL descriptor is extracted at the center of each tile (CI) or at the center of the camera (CC).


\section{Matching}  \label{sec:matching}
Given a street-level query image, our goal is to generate a `heat-map' of likely locations where this image has been taken. The semantic GIS map is therefore split into fix sized overlapping tiles, based on parameters such as average query image field-of-view and height of camera above ground.

As described previously, the rotation alignment between the descriptor used in the semantic map and the query image is unknown. In order to cope with that, we propose to rotate one of the descriptors in discrete rotation steps and compute the L2 distance for each step. 
This boils down to the computation of a circular correlation (or circular convolution) between blocks of the two descriptors corresponding to the same semantic concept. This can be implemented efficiently with a circulant matrix multiplication or even with a FFT using the circular convolution theorem~\cite{oppenheim1997signals}.


In our implementation, we are currently using the L2 distance between two descriptors. However, especially for the descriptor placed at the camera centre, several pooling regions will not be contained in the field of view. We therefore employ an asymmetric L2 distance where the distance contribution of pooling regions in the query descriptor which are not within the field of view of the camera is set to zero. The field of view of the camera can easily be estimated given the image resolution and focal length.

\section{Hierarchical Semantic Tree Representation of a GIS Map}  \label{sec:hierarchicalsemantictree}
\begin{figure}[tb]
 \centering   
\includegraphics[width=0.45\textwidth ]{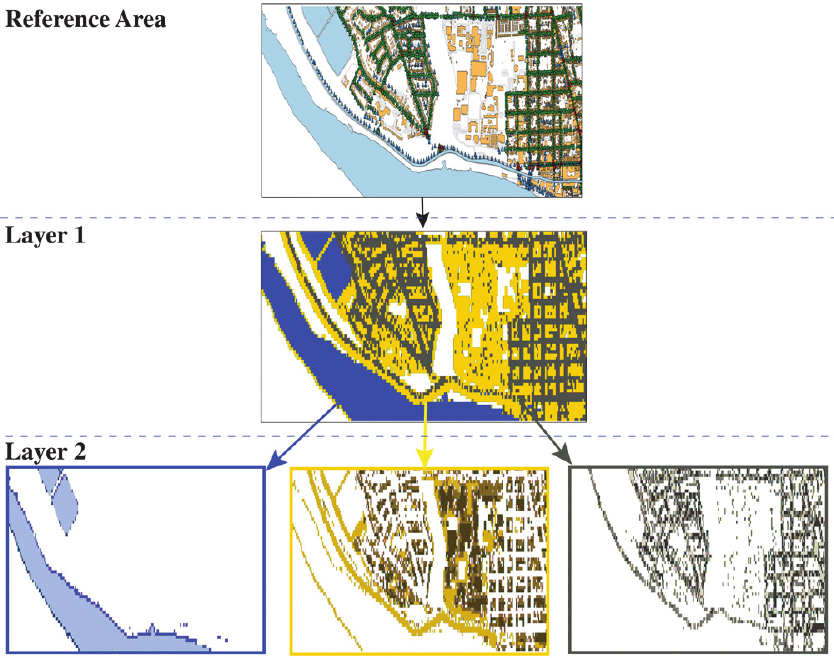}
 \caption{Two top-most layers of the semantic tree. The tiles of each layer have been split into $L = 3$ groups with the hierarchical spectral clustering, as described in the  paper. We can see how this clustering produces semantically similar clusters. Note that the 'white' areas in the top-most layer denote tiles which are mostly empty, i.e. had no semantic entries in the GIS map.}
 \label{fig:tree}
\end{figure}
The reference area covered by GIS maps is often very broad. This leads to a large number of reference tiles which the query descriptor should be compared against. Several techniques, such as k-means or kd-tree, have developed for fast nearest neighbor search~\cite{Sivic2003,Nister,Jegou2012} which we can employ to speed up this process. 
We describe a procedure inspired by k-means trees~\cite{Nister} suitable for pre-computing a hierarchical semantic tree which arranges the tiles of the map in a semantically and spatially meaningful way. This speeds up the matching and also enables fast semantic queries in a GIS system.

Our tree construction is based on hierarchical spectral clustering with $L$ branches on each level. The similarity matrix required for spectral clustering is composed of the previously described asymmetric distance between pairs of tiles of the GIS map. This provides us with a $N \times N$ similarity matrix, where $N$ denotes the number of tiles. We use spectral clustering, rather than k-means as used in k-means trees or similar methods, since we are employing our own customized distance function (asymmetric L2) while those methods often assume a standard distance.
\figref{fig:tree} shows the two top layers of the hierarchical tree ($L=3$) obtained by applying the procedure just defined on a large GIS image. What we observe is that the area has been partitioned into three well-defined semantic concepts: water (in blue), area scarcely populated (yellow) and densely populated area (grey) in the first layer. In the second layer, each of those three areas are decomposed into three other areas, and construction of the tree continues as such. The tree being semantically meaningful is a byproduct of the fact that our descriptor is targeted towards capturing semantic information.

At query time, we start traversing the tree at the root. The query image is matched against a random subset of tiles contained in each cluster of the current level. The most promising child node out of the $L$ children at each level, which represents the cluster, is found in this way. We use this technique since the cluster center is not straightforward to define for our customized distance functions. The tree is traversed all the way down to the leaf nodes to find the final match\footnote{Spilling could be introduced, as well, where more than just the most promising child is explored.}. The tree depth is $\log_L(N)$, which leads to a speed-up of roughly $\frac{N}{M \log_L(N)}$ where $M$ denotes the cardinality of the random subset of tiles explored at each level. 


\section{Experiments and Results}
\begin{figure}[tb]
 \centering   
\includegraphics[width=0.49\textwidth,keepaspectratio=true,clip=true, trim=0cm 0cm 0cm 4cm]{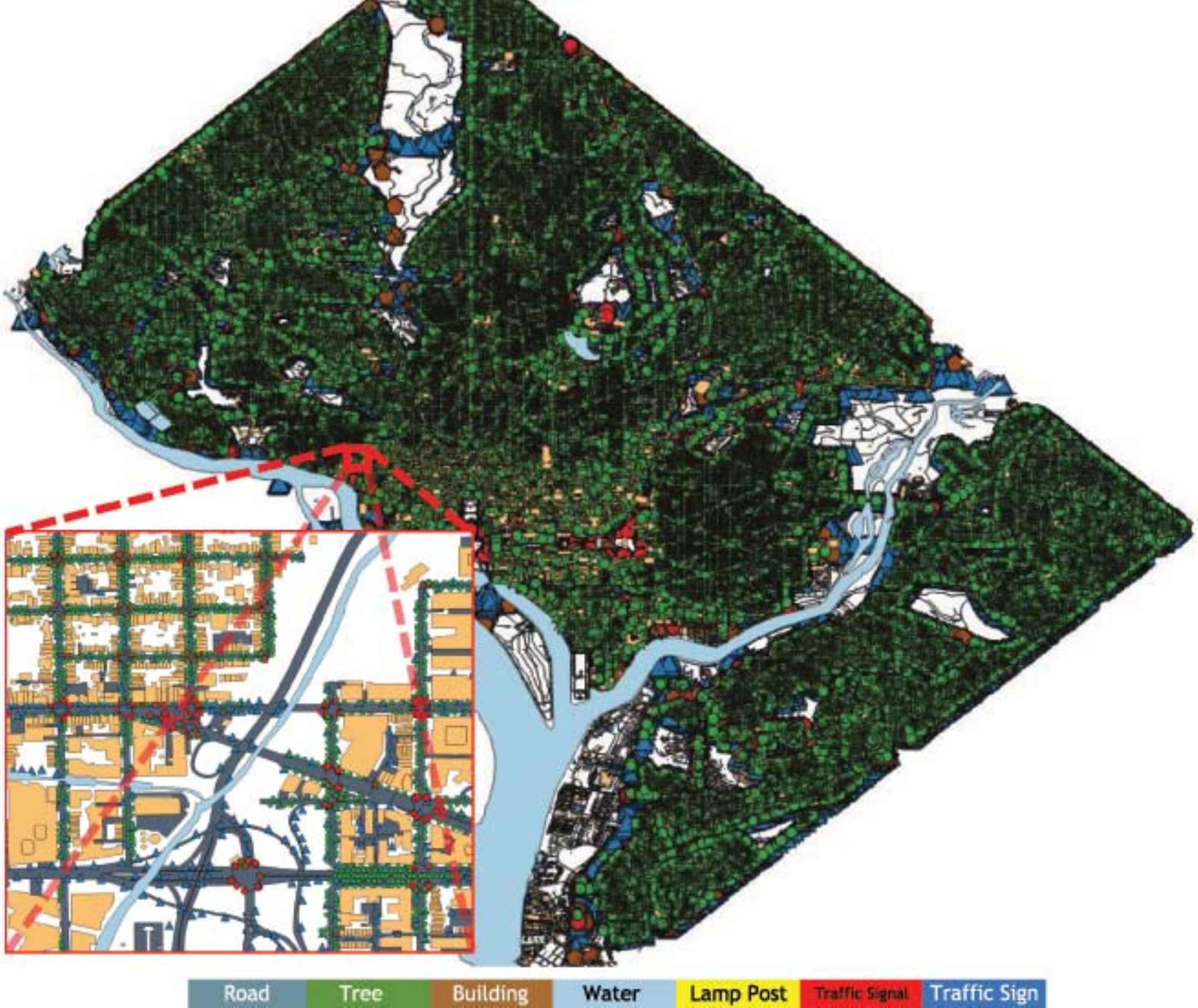}
 \caption{The area on which the cross-view matching has been tested is located in the District of Columbia, US.  On the bottom   we report the color coding for the semantic classes used in the experiments. }
  \label{fig:exp1}
\end{figure}
Our framework has been tested for the geolocalization of generic outdoor images taken in the entire ($\sim159 km^2$) of the District of Columbia, US, as extensive GIS databases of this area are made available to public~\cite{DCGIS}. The area is depicted in Figure \ref{fig:exp1} and includes a variety of different regions (water, suburban, urban). We have gathered a set of $50$ geo-tagged images from Google Maps and Panoramio taken at different locations, which serves as a benchmark for our system.
%
%
We have $N_c=7$ semantic classes, that are $C=$\{\emph{Road, Tree, Building, Water, Lamp Post, Traffic Signal, Traffic Sign}\}, and in the following experiments we are going to use different sets of classes.  The size of a tile in the GIS is around $\sim 30m^2$. We set the number of rings in our descriptors to $1$, and the number of pooling regions to $8$. 
We assume that the focal length is approximately known\footnote{A good estimate of the focal length is often available in the EXIF-header. If not, then the focal length can also be estimated from three orthogonal vanishing points, for example.}. The homography which rectifies the ground plane is then determined by the vertical vanishing point or the horizon line. 
We assume that the y-axis of the image is roughly aligned with the vertical direction and the vertical vanishing point is then given by the MLE including all the lines segments $\pm 20$~degrees w.r.t. the y-axis, see also \cite{Hartley2004}. As ground plane estimation is not the topic of our paper, we manually check and correct highly inaccurate estimates of the vanishing point in our benchmark images in order not to bias our evaluation toward mistakes in the ground plane estimation (this is done for all baselines).
Since the semantic GIS map uses metric units and the rectified image is in pixels, the size or scale of the pooling regions need to be converted between those two units. A reasonable assumption for street-level images is that the camera is roughly at $d = 1.7m$ above ground. The conversion factor between metric units and pixels is then given by $[\text{pixels}] = \frac{f}{d} [\text{meters}]$. The scale of the pooling regions is chosen such that a SSL descriptor captures significant contributions from segments within roughly $30$m. 
\begin{figure*}[tb]
 \centering   
\includegraphics[width=1\textwidth]{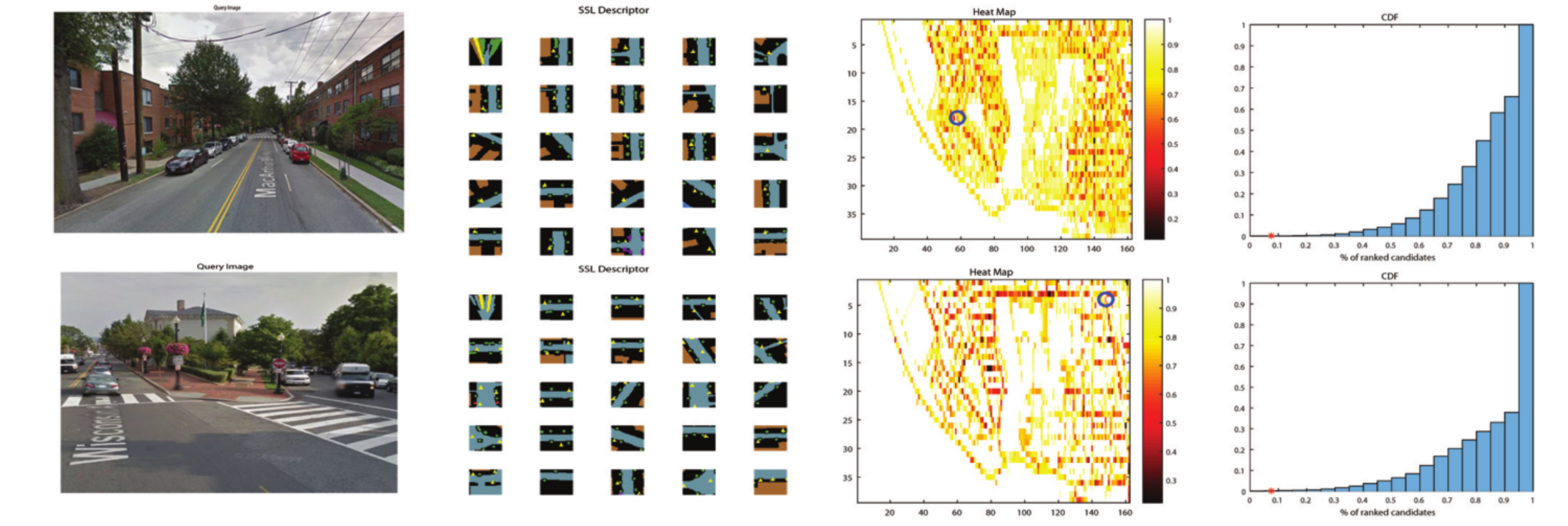}
 \caption{
 Qualitative results: Each row shows a separate query, the query image is shown in the first column.
 The second column shows the segmented and rectified query image in the top left and the top-15 tiles returned by our system (from left to right and top to bottom). 
The third column shows the ground truth location (blue circle) on top of a heat map visualization of the matching scores between the query image and the GIS map tiles (color-coded in log-scale). It can be seen that semantically similar tiles have a higher score and are therefore ranked higher. This ranking is visualized in the fourth column: each plot shows the empirical cumulative-distribution-function (CDF) of the (normalized) score between the query and all the tiles. The red star denotes the bin which contains the ground truth tile.
}
 \label{fig:quali1}
\end{figure*}
\begin{figure}[tb]
 \centering   
\includegraphics[width=0.4\textwidth]{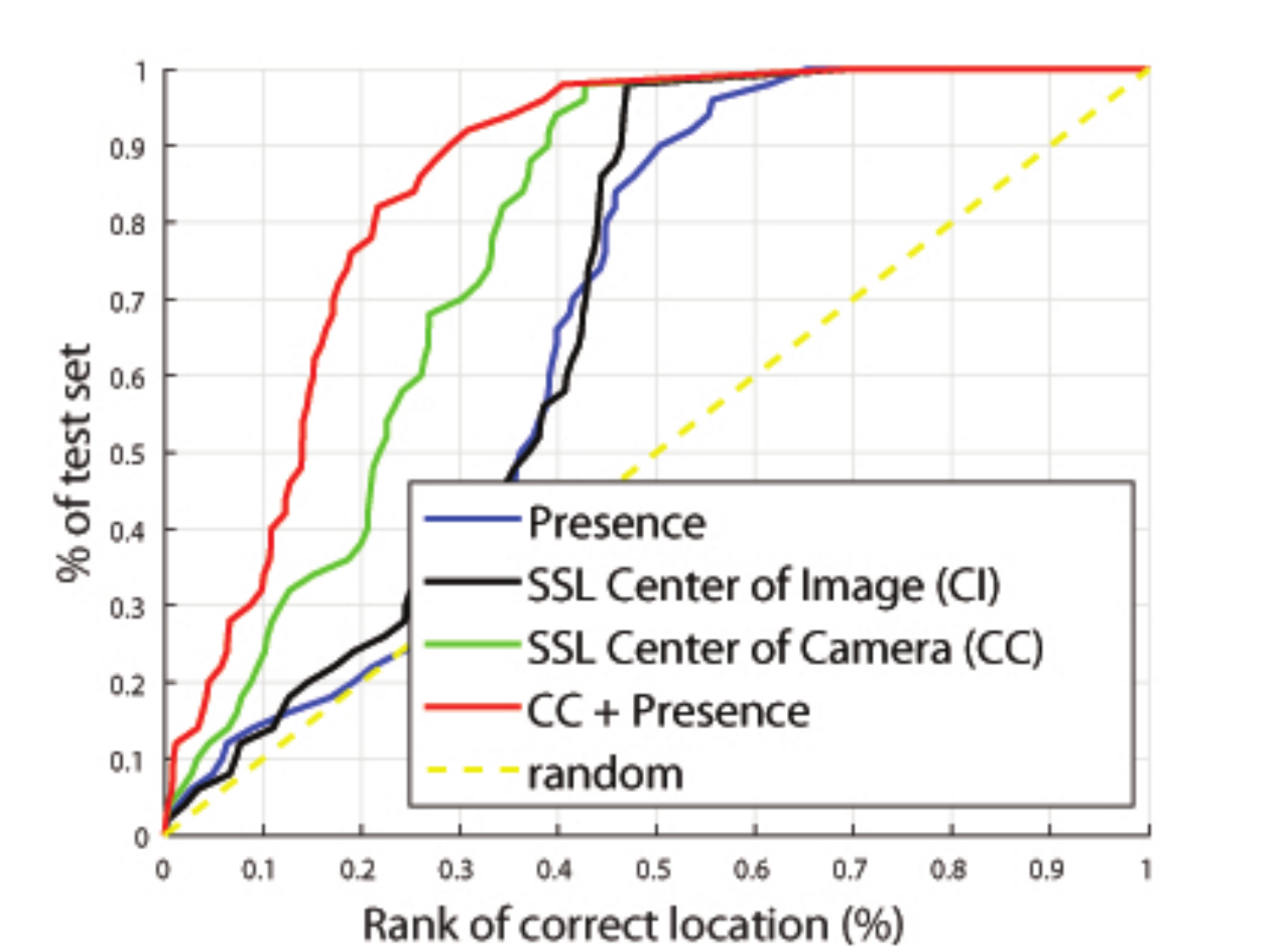}
 \caption{
Quantitative results: Analogously to the visualization in \cite{lin2013cross}, this figure plots the percentage of queries which contained the ground truth tile in a short-list of the best scoring tiles against the normalized size of that short-list (i.e. the x-axis denotes the fraction of tiles contained in the short-list). The red curve shows how the combination of SSL descriptor in the center of camera and Presence term outperforms the other methods. \vspace{-5pt}}
 \label{fig:quant1}
\end{figure}

\figref{fig:quali1} reports detailed qualitative results of our proposed cross-view matching scheme for several sample queries. Several interesting observations can be made in this figure. First, as the covered area is very large, even such generic semantic cues can narrow down the search space to often $<5\%$. Second, the semantic and geometric similarity among the top 15 matching GIS tiles shows the proposed method is successfully capturing such properties and is yet forgiving of the modeled uncertainties by not being overly discriminative; note that the ground truth location (marked in the CDF) is often among the top few percents. Third, the heat map correlates well with the semantic content of the image. 

As for quantitative results, we compare different Nearest Neighbor (NN) classifiers with the following feature vectors: SSL descriptor in the center of image (per query image/GIS tile), SSL descriptor in the center of the camera, binary indicator vector encoding the presence or absence of semantic concepts (\emph{Presence} term), SSL descriptor in the center of the camera plus the Presence term. The last method is random matching. The superior results of the SSL plus Presence matching reported in \figref{fig:quant1} confirm the necessity of jointly using semantic and coarse geometry for a successful localization task.
We also suspect that placing the SSL descriptor in the center of camera results in better performance than placing the descriptor in the center of the image because the former approach is less sensible to tiling quantization artifacts: e.g. objects on the left side of the field of view will generally remain on the left side even when the viewpoint is moved to the closest tile location.

\begin{figure}[tb]
\centering   
\includegraphics[width=0.4\textwidth]{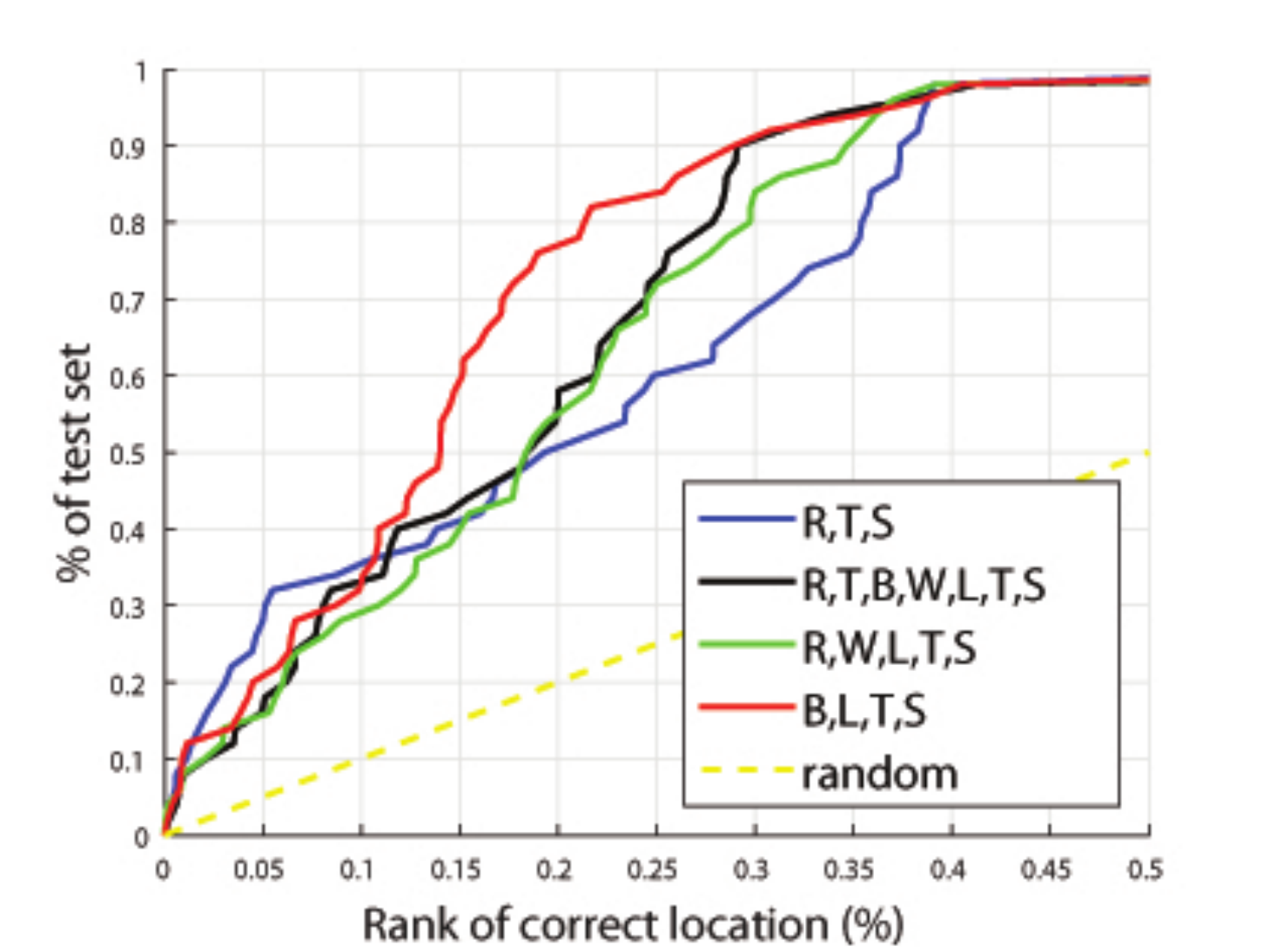}
\caption{ Quantification of the impact of different  semantic configurations (in the legend R is road, T is tree, B is building, W is water, L is lamp post, T is traffic signal, S is traffic sign). \vspace{-3pt}}
\label{fig:excl}
\end{figure}

We used the subset $C^s=$\{\emph{Building, Lamp Post, Traffic Signal, Traffic Sign}\} of the $N_c$ semantic classes since this subset yielded the best overall geo-localization result. 
For further evaluation, \figref{fig:excl} depicts the results of SSL plus Presence matching over four different sets of semantic classes, which shows the contribution of each semantic class in the overall geo-localization results.
Interestingly, the curves also show how certain sets of semantic classes may even mislead the geo-localization process. We believe this is due to un-informativeness (e.g. being too common) of some classes as well as several sources of noise whose magnitude can vary between different classes (e.g. inaccurate entries in GIS map, semantic segmentation misclassifications, etc.).
If enough training data were available, appropriate weights could be learned for each class.

\section{Summary}

This paper proposed an approach for cross-view matching between a street-level image and a GIS map. This problem was addressed in a semantic way. A fast Semantic Segment Layout descriptor has been proposed which jointly captures the presence of segments with a certain semantic concept and the spatial layout of those segments. As our experimental evaluation showed, this enabled matching a street-level image to a large reference map based on purely semantic cues of the scene and their coarse spatial layout. The results confirm that the semantic and topological cues captured by our method significantly narrow down the search area. This can be used as an effective pre-processing for other less efficient but more accurate localization techniques, such as street view based methods. 
%
%

\textbf{Acknowledgments: }
This work has been supported by the Max Planck Center for Visual Computing and Communication. We also acknowledge the support of NSF CAREER grant (N1054127).



\begin{thebibliography}{10}\itemsep=-1pt

\bibitem{DCGIS}
{Washington DC Open GIS Data}.
\newblock \url{http://opendata.dc.gov/}.

\bibitem{Abou}
K.~T. Abou–Moustafa and F.~P. Ferrie.
\newblock A note on metric properties of some divergence measures: The gaussian
  case.
\newblock In {\em ACML, JMLR W\&CP}, 2012.

\bibitem{ardeshir2014gis}
S.~Ardeshir, A.~R. Zamir, A.~Torroella, and M.~Shah.
\newblock Gis-assisted object detection and geospatial localization.
\newblock In {\em ECCV}, pages 602--617. Springer, 2014.

\bibitem{baatz2012leveraging}
G.~Baatz, K.~K{\"o}ser, D.~Chen, R.~Grzeszczuk, and M.~Pollefeys.
\newblock Leveraging 3d city models for rotation invariant place-of-interest
  recognition.
\newblock {\em IJCV}, 96(3):315--334, 2012.

\bibitem{baatz2012large}
G.~Baatz, O.~Saurer, K.~K{\"o}ser, and M.~Pollefeys.
\newblock Large scale visual geo-localization of images in mountainous terrain.
\newblock In {\em ECCV}, pages 517--530. Springer, 2012.

\bibitem{bansal2012ultra}
M.~Bansal, K.~Daniilidis, and H.~Sawhney.
\newblock Ultra-wide baseline facade matching for geo-localization.
\newblock In {\em Computer Vision--ECCV 2012. Workshops and Demonstrations},
  pages 175--186. Springer, 2012.

\bibitem{bay2005wide}
H.~Bay, V.~Ferrari, and L.~Van~Gool.
\newblock Wide-baseline stereo matching with line segments.
\newblock In {\em CVPR}, pages 329--336. IEEE, 2005.

\bibitem{bazin2012globally}
J.-C. Bazin, Y.~Seo, C.~Demonceaux, P.~Vasseur, K.~Ikeuchi, I.~Kweon, and
  M.~Pollefeys.
\newblock Globally optimal line clustering and vanishing point estimation in
  manhattan world.
\newblock In {\em CVPR}. IEEE, 2012.

\bibitem{belongie2000shape}
S.~Belongie, J.~Malik, and J.~Puzicha.
\newblock Shape context: A new descriptor for shape matching and object
  recognition.
\newblock In {\em NIPS}, volume~2, page~3, 2000.

\bibitem{denis2008efficient}
P.~Denis, J.~H. Elder, and F.~J. Estrada.
\newblock Efficient edge-based methods for estimating manhattan frames in urban
  imagery.
\newblock In {\em ECCV}, pages 197--210. Springer, 2008.

\bibitem{durand2014semantic}
T.~Durand, D.~Picard, N.~Thome, and M.~Cord.
\newblock Semantic pooling for image categorizatin using multiple kernel
  learning.
\newblock In {\em ICIP}, 2014.

\bibitem{fischler1981random}
M.~A. Fischler and R.~C. Bolles.
\newblock Random sample consensus: a paradigm for model fitting with
  applications to image analysis and automated cartography.
\newblock {\em Communications of the ACM}, 24(6):381--395, 1981.

\bibitem{frueh2003constructing}
C.~Frueh and A.~Zakhor.
\newblock Constructing 3d city models by merging ground-based and airborne
  views.
\newblock In {\em CVPR}, volume~2, pages 554--562. IEEE, 2003.

\bibitem{Hartley2004}
R.~Hartley and A.~Zisserman.
\newblock {\em Multiple view geometry in computer vision}.
\newblock Cambridge University Press, 2nd edition, 2004.

\bibitem{hoiem2008putting}
D.~Hoiem, A.~A. Efros, and M.~Hebert.
\newblock Putting objects in perspective.
\newblock {\em IJCV}, 80(1):3--15, 2008.

\bibitem{Jegou2012}
H.~J\'{e}gou, F.~Perronnin, M.~Douze, J.~Sanchez, P.~Perez, and C.~Schmid.
\newblock {Aggregating Local Image Descriptors into Compact Codes}.
\newblock {\em PAMI}, 34(9), 2012.

\bibitem{kaminsky2009alignment}
R.~S. Kaminsky, N.~Snavely, S.~M. Seitz, and R.~Szeliski.
\newblock Alignment of 3d point clouds to overhead images.
\newblock In {\em CVPR Workshop}, pages 63--70. IEEE, 2009.

\bibitem{lin2013cross}
T.-Y. Lin, S.~Belongie, and J.~Hays.
\newblock Cross-view image geolocalization.
\newblock In {\em CVPR}, pages 891--898. IEEE, 2013.

\bibitem{lowe2004}
D.~Lowe.
\newblock Distinctive image features from scale-invariant keypoints.
\newblock {\em IJCV}, 60(2), 2004.

\bibitem{malisiewicz2011ensemble}
T.~Malisiewicz, A.~Gupta, and A.~A. Efros.
\newblock Ensemble of exemplar-svms for object detection and beyond.
\newblock In {\em ICCV}. IEEE, 2011.

\bibitem{matas2004robust}
J.~Matas, O.~Chum, M.~Urban, and T.~Pajdla.
\newblock Robust wide-baseline stereo from maximally stable extremal regions.
\newblock {\em Image and vision comp.}, 22(10):761--767, 2004.

\bibitem{Nister}
D.~Nister and H.~Stewenius.
\newblock Scalable recognition with a vocabulary tree.
\newblock In {\em CVPR}, 2006.

\bibitem{oppenheim1997signals}
A.~V. Oppenheim and A.~S. Willsky.
\newblock {\em Signals and systems}.
\newblock Prentice-Hall, 1997.

\bibitem{ramalingam2010skyline2gps}
S.~Ramalingam, S.~Bouaziz, P.~F. Sturm, and M.~Brand.
\newblock Skyline2gps: Localization in urban canyons using omni-skylines.
\newblock In {\em IROS}, pages 3816--3823, 2010.

\bibitem{ren2012rgb}
X.~Ren, L.~Bo, and D.~Fox.
\newblock Rgb-(d) scene labeling: Features and algorithms.
\newblock In {\em CVPR}. IEEE, 2012.

\bibitem{saxena2007learning}
A.~Saxena, M.~Sun, and A.~Y. Ng.
\newblock Learning 3-d scene structure from a single still image.
\newblock In {\em ICCV}. IEEE, 2007.

\bibitem{Sfikas2005}
G.~Sfikas, C.~Constantinopoulos, A.~Likas, and N.~P. Galatsanos.
\newblock An analytic distance metric for gaussian mixture models with
  application in image retrieval.
\newblock In {\em ANN}, pages 835--840. Springer, 2005.

\bibitem{shan2014accurate}
Q.~Shan, C.~Wu, B.~Curless, Y.~Furukawa, C.~Hernandez, and S.~M. Seitz.
\newblock Accurate geo-registration by ground-to-aerial image matching.
\newblock In {\em 3DV}. IEEE, 2014.

\bibitem{Sivic2003}
J.~Sivic and A.~Zisserman.
\newblock Video google: A text retrieval approach to object matching in videos.
\newblock In {\em ICCV}, 2003.

\bibitem{tola2008fast}
E.~Tola, V.~Lepetit, and P.~Fua.
\newblock A fast local descriptor for dense matching.
\newblock In {\em CVPR}, pages 1--8. IEEE, 2008.

\bibitem{verhagen2014scale}
B.~Verhagen, R.~Timofte, and L.~Van~Gool.
\newblock Scale-invariant line descriptors for wide baseline matching.
\newblock In {\em WACV}, pages 493--500. IEEE, 2014.

\end{thebibliography}
\end{document}